\documentclass{article}
\usepackage{iclr2026_conference,times}
\usepackage{graphicx}
\graphicspath{{figures/}{../figures/}}
\usepackage{placeins}
\usepackage{url}
\usepackage{xurl}
\usepackage{wrapfig}
\usepackage{hyperref}
\usepackage{tikz}
\usetikzlibrary{arrows.meta,positioning}

\usepackage{amsmath}
\usepackage{amssymb}

\usepackage{booktabs}
\usepackage{adjustbox}

\title{Proteus: A Truncation-Robust Entropy Model for Progressive LiDAR Compression}

\iclrfinalcopy 

\author{
  Yihan Qiu$^{1,2}$,~Xiaodong Lin$^{2}$,~Baoquan Zhao$^{2}$,~Hailong Jiao$^{1}$,~Ge Li$^{1}$\thanks{Corresponding author.} \\
  $^{1}$Peking University \qquad\qquad\qquad $^{2}$Sun Yat-sen University \\
  \texttt{qiuyh33@gmail.com} \qquad \texttt{\{jiaohailong,~geli\}@pku.edu.cn} \\
  \texttt{\{linxd58@mail2,~zhaobaoquan@mail\}.sysu.edu.cn}
}

\begin{document}
\raggedbottom
\maketitle

\begin{abstract}
LiDAR point clouds provide explicit, deterministic physical boundaries critical for collaborative safety-critical perception. However, wireless channels inherently impair and corrupt transmitted signals. Existing robust frameworks (such as deep JSCC or MDC) attempt to counter these channel impairments through statistical or parametric estimation, turning exact physical measurements into unverified algorithmic estimates. To address this, we propose Proteus, a learned LiDAR codec operating on 2D range images. By decoupling the frame representation into independent coders for the \textbf{sig}nificant range bit-planes (SIG) and the \textbf{ins}ignificant range bit-planes and attributes (INS), Proteus achieves overall stream-level truncation robustness. The non-truncatable SIG block encodes the most significant range bit-planes to establish a necessary, self-contained perceptual lower bound, below which the reconstructed point cloud is severely degraded. Meanwhile, INS employs bit-plane slicing representation and coding, ensuring that range truncation mathematically maps to a deterministic spatial precision degradation. Subordinate attributes are reconstructed via a hybrid lossless-predictive method, leveraging the decoded geometry as a strong structural prior for fine-grained approximation. Furthermore, strategic ordering within INS prioritizes geometry over attributes under bandwidth drops. Experimental results on the Waymo Open Dataset and SemanticKITTI demonstrate that Proteus tolerates up to approximately 70\% bitstream truncation, while outperforming established standards (G-PCC, Draco, and JPEG XL) and the representative learned compressor Unicorn under ideal channel conditions.
\end{abstract}


\section{Introduction}
\label{sec:introduction}

LiDAR point clouds are a key data structure for scene understanding in autonomous driving, robotics, and remote sensing, providing rich geometric, shape, and scale information \citep{guo2021deep}. Beyond geometry, modern LiDARs also record per-point attributes such as reflectance intensity that carry complementary cues for downstream perception \citep{zhu2025serlic}. Crucially, active LiDAR sensors directly measure explicit, deterministic physical boundaries \citep{royo2019lidar}. In safety-critical tasks such as collision detection, freespace boundary delineation, and emergency hazard avoidance, these precise physical measurements serve as an indispensable and mathematically verifiable safety net \citep{royo2019lidar,mao20233dobject}. While single-agent perception is inherently limited by physical line-of-sight constraints and severe occlusions \citep{wang2020v2vnet}, collaborative applications resolve these bottlenecks by continuously sharing these deterministic physical boundaries. Yet, in collaborative applications such as vehicle-to-everything (V2X) communications or UAV swarms, transmitting these dense, high-volume point clouds \citep{gao2025deeppccsurvey} over shared and highly dynamic wireless channels \citep{hu2022where2comm} places both high-efficiency compression and channel-robust transmission on the same critical path.

Specifically, under modern communication protocol stacks \citep{3gpp2024ts23287}, the wireless channel between agents is primarily subject to bandwidth fluctuation \citep{hu2022where2comm} and transient link outages \citep{li2023lossyv2v,ren2024v2xincop}. Bandwidth fluctuation causes in-transit stream truncation \citep{jeon2023ctc}. However, conventional codecs collapse under such truncation, requiring latency-heavy feedback loops to renegotiate the bitrate and stalling the real-time pipeline \citep{li2024palette}. This demands a prefix-decodable bitstream that yields a valid reconstruction without prior coordination. Meanwhile, transient link outages cause complete, frame-level loss, requiring each frame to be self-contained, as inter-frame prediction breaks down when reference frames are lost in transit \citep{wiegand2003h264,zhou2022riddlelidar}. Most existing LiDAR codecs, primarily designed for ideal channels, provide neither property \citep{gao2025deeppccsurvey,wang2025aversatile_geom,you2025renoreal}.

To mitigate these transmission bottlenecks, recent learned robust communication frameworks, such as deep joint source-channel coding (Deep JSCC) \citep{bourtsoulatze2018deepjoint,zhang2024pcst} or multiple description coding (MDC) \citep{elgamal1982achievable,chen2024mdc}, have successfully achieved continuous quality degradation under channel impairments, maintaining competitive PSNR or downstream mAP. However, because these approaches primarily rely on statistical or parametric estimation to reconstruct the missing or corrupted geometric regions, the decoded coordinates inevitably become an algorithmic estimate rather than an exact, faithful measurement \citep{zhang2024pcst,cheng2024grace,hung2022errorconcealment}. For safety-critical decisions, preserving the exact, physically bounded nature of the sensor's measurement, even at a coarser resolution, remains the indispensable bedrock of trustworthy collaborative intelligence. This prompts the question: \textbf{\emph{how can a learned codec achieve robust progressive compression while strictly preserving mathematically bounded physical boundaries?}}

To address these challenges, we propose Proteus, a learned LiDAR codec operating on 2D range images. By restricting coding dependencies strictly intra-frame, Proteus achieves frame-level packet-loss resistance. To resolve truncation vulnerability, we decouple the single-frame representation into two independent coders: a self-contained bits-back coder for the \textbf{sig}nificant range bit-planes (SIG), and a FIFO range coder for the \textbf{ins}ignificant range bit-planes and attributes (INS). The non-truncatable SIG block secures the fundamental geometric structure, without which the reconstructed point cloud loses its perceptual utility, while utilizing an Autoregressive Initial Bits (ArIB) mechanism \citep{ryder2022splithierarchical,zhang2024learnedlossless} to eliminate the initial state bits overhead inherent in single-frame bits-back decoding. Meanwhile, INS employs bit-plane slicing representation and coding, ensuring that range truncation mathematically maps to a deterministic spatial precision degradation of physical distances, rather than relying on unconstrained statistical estimation. Furthermore, strategic stream serialization within INS prioritizes geometric structures over subordinate attributes under bandwidth dips; when attributes are truncated, they are reconstructed via a hybrid lossless-predictive approach using the decoded geometry as a prior to mitigate quality loss. Since SIG accounts for roughly 30\% of the bitstream, Proteus remains decodable under up to 70\% truncation.

Our contributions are as follows.

\begin{itemize}
\item (i) We address the fundamental incompatibility between truncation robustness and high-efficiency bits-back coding equipped with the Autoregressive Initial Bits (ArIB) mechanism to eliminate single-frame overhead. This enables progressive, graceful degradation under truncation with only a minor penalty in lossless compression efficiency.
\item (ii) We design a joint geometry-attribute coding framework that strategically orders subordinate attribute data after geometry in the truncatable stream. This allows the system to naturally sacrifice attribute precision first during bandwidth dips, maintaining respectable downstream perception performance even when up to 70\% of the overall bitstream is truncated.
\item (iii) Our proposed framework achieves competitive compression efficiency compared to established standards and industrial codecs (such as G-PCC, Draco, and JPEG XL) and the representative learned baseline Unicorn, achieving superior compression efficiency while offering remarkable robustness under channel degradation.
\end{itemize}

\section{Related Work}
\label{sec:related_work}

\subsection{Point Cloud Compression}

A point cloud consists of point positions and per-point attributes such as reflectance \citep{gao2025deeppccsurvey,fang20223dac,zhu2025serlic}, and most compression pipelines encode these two streams in separate stages. Geometry coders organize the unordered point set into a structured representation amenable to entropy coding \citep{liu2026nextbit}. Octree-based methods represent geometry as occupancy trees, where G-PCC \citep{graziosi2020anoverview} employs hand-designed contexts and learned variants \citep{huang2021octsqueezeoctree,fu2022octattentionoctree,song2023efficienthierarchical} train neural networks to predict node occupancy distributions. Voxel approaches discretize 3D space into a regular grid and predict per-cell occupancy with 3D convolutions \citep{he2022densitypreserving,que2021voxelcontext}. Sparse tensor methods \citep{wang2022sparsepcgc,wang2025aversatile_geom,you2025renoreal} apply sparse convolutions only at occupied voxels, which lowers the cost of voxel-based processing for sparse LiDAR scans. Alternatively, range images offer a memory-efficient data structure to organize and represent point clouds \citep{meyer2019lasernetan,wang2021rpcc,liu2023birdpcc}.

To exploit temporal redundancy, several coders condition each frame on previously decoded frames, for example through K-nearest-neighbor lookup in the previous scan \citep{biswas2020muscle,wang2025aversatile_geom,zhou2022riddlelidar}, range-view optical flow \citep{song2025lowlatency}, or explicit key-frame / predicted-frame structures with motion compensation \citep{feng2020spatiotemporal,fan2022ddpcc}. These inter-frame methods improve rate-distortion under an ideal channel but rely on reference-frame availability: a lost reference propagates errors to subsequent frames, and the first frame of a sequence lacks a reference \citep{wiegand2003h264,zhou2022riddlelidar}.

For attribute coding, the G-PCC standard \citep{graziosi2020anoverview} is built on fixed transforms, notably RAHT and the predicting transform, while recent learned coders replace them with neural transforms and context models \citep{wang2025aversatile_attr,fang20223dac,zhu2025serlic}. However, the common practice of running separate networks for geometry and attributes \citep{wang2025aversatile_attr,wang2025aversatile_geom} increases end-to-end latency \citep{song2025lowlatency}.

\subsection{Robust Compression and Transmission}

Robust transmission under bandwidth fluctuation or packet loss has received less attention than rate-distortion optimization, particularly for point clouds \citep{gao2025deeppccsurvey,zhang2024pcst,chen2024mdc}. We group existing approaches by their robustness mechanism into three families.

The first family abandons source-channel separation, directly mapping the signal to channel symbols via a learned encoder-decoder. Without explicit source coding, the decoder reconstructs the source directly from noisy channel outputs, allowing quality to degrade continuously without the cliff effect. The idea originates in pseudo-analog video \citep{jakubczak2010softcast}, is revived by deep JSCC for images \citep{bourtsoulatze2018deepjoint}, with recent variants adopting explicit generative models to trade pixel fidelity for perceptual quality \citep{erdemir2023generativejscc}, and is extended to point clouds \citep{zhang2024pcst}.

The second family retains conventional digital source coding and addresses packet loss \citep{hung2022errorconcealment}. Partitioning approaches split the bitstream into independently decodable parts, grounded in the information-theoretic multiple-description framework \citep{elgamal1982achievable} and instantiated in industrial video error-resilience tools \citep{wiegand2003h264} and point-cloud multiple description coding \citep{chen2024mdc}. Alternatively, learning-based methods train the codec under random masking, enabling recovery from partial bitstreams without explicit partitioning or post-hoc concealment \citep{cheng2024grace}. In both branches, regions whose packets are lost are reconstructed by the codec rather than preserved from the source, so the decoded signal there is an algorithmic estimate \citep{cheng2024grace,hung2022errorconcealment}.

The third family encodes the bitstream so that any prefix decodes to a valid lower-quality reconstruction \citep{liu2026nextbit,jeon2023ctc}. The embedded-bitstream principle, pioneered by zerotree wavelets \citep{shapiro1993ezw} and JPEG 2000 \citep{skodras2001jpeg2000}, requires no negotiation or learned decoders. It ensures deterministic degradation, where any truncated reconstruction is a mathematically bounded, non-parametric projection of the source. Conversely, neural latent-variable schemes like PLONQ \citep{lu2021plonq} and DPICT \citep{lee2022dpict} rely on parametric synthesis, using learned statistical estimation to reconstruct truncated codes rather than deterministic degradation. Integrating the principle of deterministic progressive degradation with neural latent-variable frameworks remains an open challenge.

\section{Preliminaries}
\label{sec:preliminaries}

\subsection{Range Image Representation}

A LiDAR sensor scans the scene by emitting $H \times W$ laser shots along $H$ elevation angles $\theta = \{\theta_1, \ldots, \theta_H\}$ and $W$ azimuth angles $\phi = \{\phi_1, \ldots, \phi_W\}$. Each shot returns a scalar range value $r_{i,j}$ \citep{royo2019lidar} that measures the radial distance to the first reflective surface along beam $(i, j)$. The $H \times W$ grid of range values, optionally accompanied by per-pixel attributes such as reflectance intensity, is the range image, a 2D intermediate representation aligned with the LiDAR scanning pattern \citep{milioto2019rangenetfast}. The Cartesian coordinates of the corresponding point are recovered losslessly from the range value via

\begin{equation}
\begin{aligned}
x &= r_{i,j} \cos\theta_i \cos\phi_j, \\
y &= r_{i,j} \cos\theta_i \sin\phi_j, \\
z &= r_{i,j} \sin\theta_i,
\end{aligned}
\end{equation}

where the emission angles $\theta$ and $\phi$ are fixed by the predefined sensor scanning pattern and known a priori at the receiver \citep{sun2020waymo,zhou2022riddlelidar}. The point cloud is therefore a derived representation: transmitting the scalar range value $r$ suffices to reconstruct the 3D position, which is more efficient than transmitting the three Cartesian coordinates directly. We adopt the range image as the working representation throughout this paper.

\subsection{Bits-back Coding}

Consider a latent-variable model $p_\theta(\mathbf{x}, \mathbf{z}) = p_\theta(\mathbf{x}|\mathbf{z}) p(\mathbf{z})$ with variational posterior $q_\phi(\mathbf{z}|\mathbf{x})$. Bits-back coding \citep{townsend2019practicallossless} encodes $\mathbf{x}$ in three steps:

\begin{enumerate}
\item Decode $\mathbf{z}$ with $q_\phi(\mathbf{z}|\mathbf{x})$ from initial bits.
\item Encode $\mathbf{x}$ with $p_\theta(\mathbf{x}|\mathbf{z})$.
\item Encode $\mathbf{z}$ with $p(\mathbf{z})$.
\end{enumerate}

Decoding reverses the order: $\mathbf{z}$ is decoded under $p(\mathbf{z})$, $\mathbf{x}$ under $p_\theta(\mathbf{x}|\mathbf{z})$, and the initial bits are released by encoding $\mathbf{z}$ under $q_\phi(\mathbf{z}|\mathbf{x})$. Asymmetric Numeral Systems \citep{duda2014asymmetricnumeral} provides an entropy coder whose encoding and decoding orders are reversed, which is required by this encode-then-decode structure. The average code length over a sequence of $N$ images is

\begin{equation}
\begin{aligned}
\bar{C} ={}& \frac{1}{N} \sum_{i=1}^{N} \Bigl[
-\log p_\theta(\mathbf{x}^{(i)} \mid \mathbf{z}^{(i)})
- \log p(\mathbf{z}^{(i)}) \\
&\qquad + \log q_\phi(\mathbf{z}^{(i)} \mid \mathbf{x}^{(i)})\Bigr]
- \frac{1}{N}
\log q_\phi(\mathbf{z}^{(1)} \mid \mathbf{x}^{(1)}),
\end{aligned}
\end{equation}

where the first term is the negative ELBO and the second term is the initial-bits overhead amortized over the sequence. For a single image the overhead $-\log q_\phi(\mathbf{z}|\mathbf{x})$ is not amortized, which limits bits-back coding to dataset compression unless the overhead is supplied by another mechanism.

\subsection{ArIB and ArIB-BPS}

Autoregressive initial bits (ArIB), introduced by SHVC \citep{ryder2022splithierarchical}, partitions the data sequence into two subsets $\mathbf{x}_{1:s}$ and $\mathbf{x}_{s+1:n}$, where $n$ is the total number of subimages and $s$ is the split index. Latent variables are used exclusively for $\mathbf{x}_{1:s}$, and the bits used to encode $\mathbf{x}_{s+1:n}$ serve as the initial bits for bits-back coding on $\mathbf{x}_{1:s}$. When $\mathbf{x}_{s+1:n}$ supplies enough bits, the single-image overhead of Section 3.2 is eliminated. SHVC realizes ArIB by splitting the image along the space dimension into $n$ spatial subimages: pixels sharing the same coordinate end up adjacent in the original image, so every subimage carries comparable importance for the latent. Excluding the latent from some subimages then underuses it, and the strong autoregressive prior accumulated towards the end of the subimage sequence drives the posterior towards the prior, a failure mode known as posterior collapse. ArIB-BPS \citep{zhang2024learnedlossless} realizes ArIB by splitting along the bit-plane dimension instead, taking $n = d$ where $d$ is the bit depth. For a $d$-bit image, plane $l$ is

\begin{equation}
\mathbf{x}^l = \lfloor \mathbf{x} / 2^{d-l} \rfloor \bmod 2,
\end{equation}

and the planes are ordered from the most significant plane (MSP, $l=1$) to the least significant plane (LSP, $l=d$). The volume of global data modality decreases from MSP to LSP; since latent variables capture this global modality, their importance is hypothesized to decrease along the same order. ArIB-BPS names $\mathbf{x}^{1:s}$ the significant planes and $\mathbf{x}^{s+1:d}$ the insignificant planes; the insignificant planes are conditionally independent of the latent and available as initial bits, while the posterior-collapse risk is shifted onto the planes that least need the latent. The resulting rate decomposes as

\begin{equation}
R_{\mathbf{x}} = -\log p(\mathbf{x}^{1:s}|\mathbf{z}) - \log p(\mathbf{x}^{s+1:d}|\mathbf{x}^{1:s}).
\end{equation}

\section{Method}
\label{sec:method}

\begin{figure}[t]
\centering
\begin{minipage}[c]{0.62\textwidth}
\centering
\includegraphics[trim=88pt 272pt 336pt 92pt,clip,width=\linewidth]{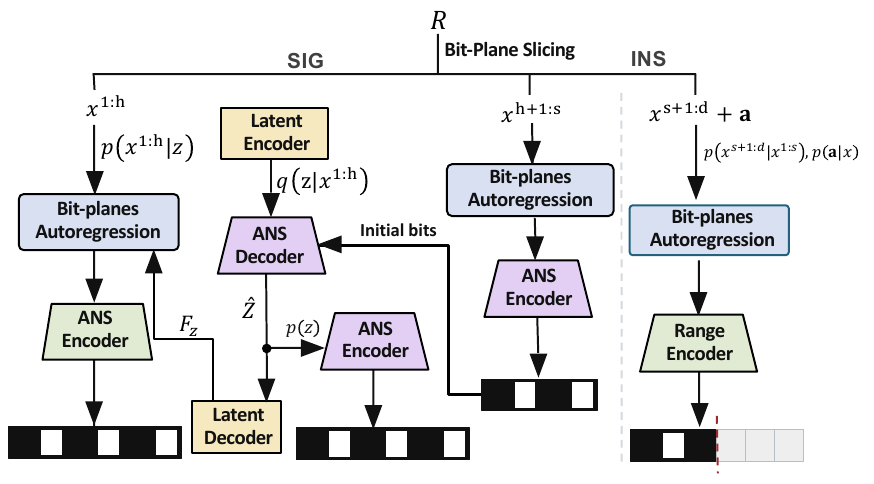}\\[-2pt]
{\small (a) Decoupled SIG (bits-back) and INS (range coding) pipelines}
\end{minipage}
\hfill
\begin{minipage}[c]{0.36\textwidth}
\centering
\includegraphics[trim=270pt 290pt 357pt 125pt,clip,width=\linewidth]{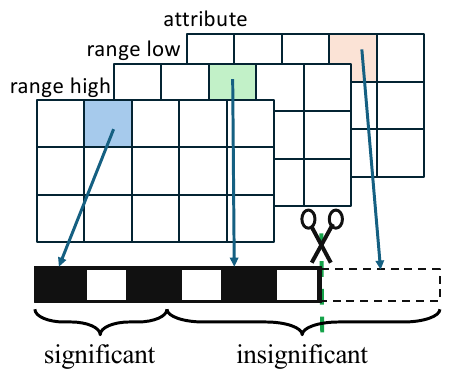}\\[-2pt]
{\small (b) Truncatable bitstream}
\end{minipage}
\caption{Overview of the Proteus framework designed for truncation-robust LiDAR compression.}
\label{fig:overview}
\end{figure}

\begin{figure}[t]
\centering
\begin{minipage}[t]{0.32\textwidth}
\centering
\vspace{0pt}
\includegraphics[width=\linewidth]{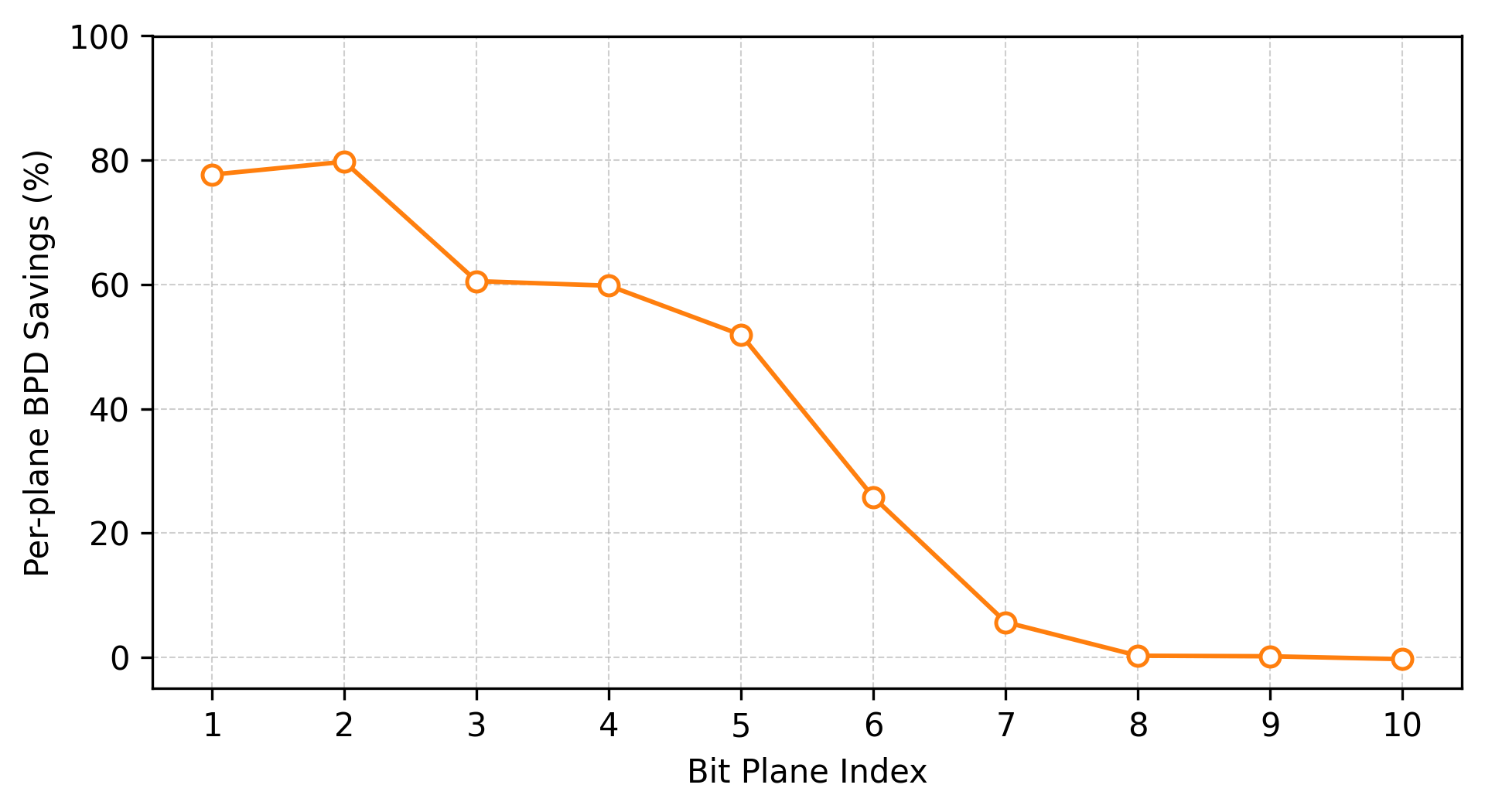}\\[-2pt]
{\scriptsize (a) SemanticKITTI BPD savings}
\end{minipage}
\hfill
\begin{minipage}[t]{0.32\textwidth}
\centering
\vspace{0pt}
\includegraphics[width=\linewidth]{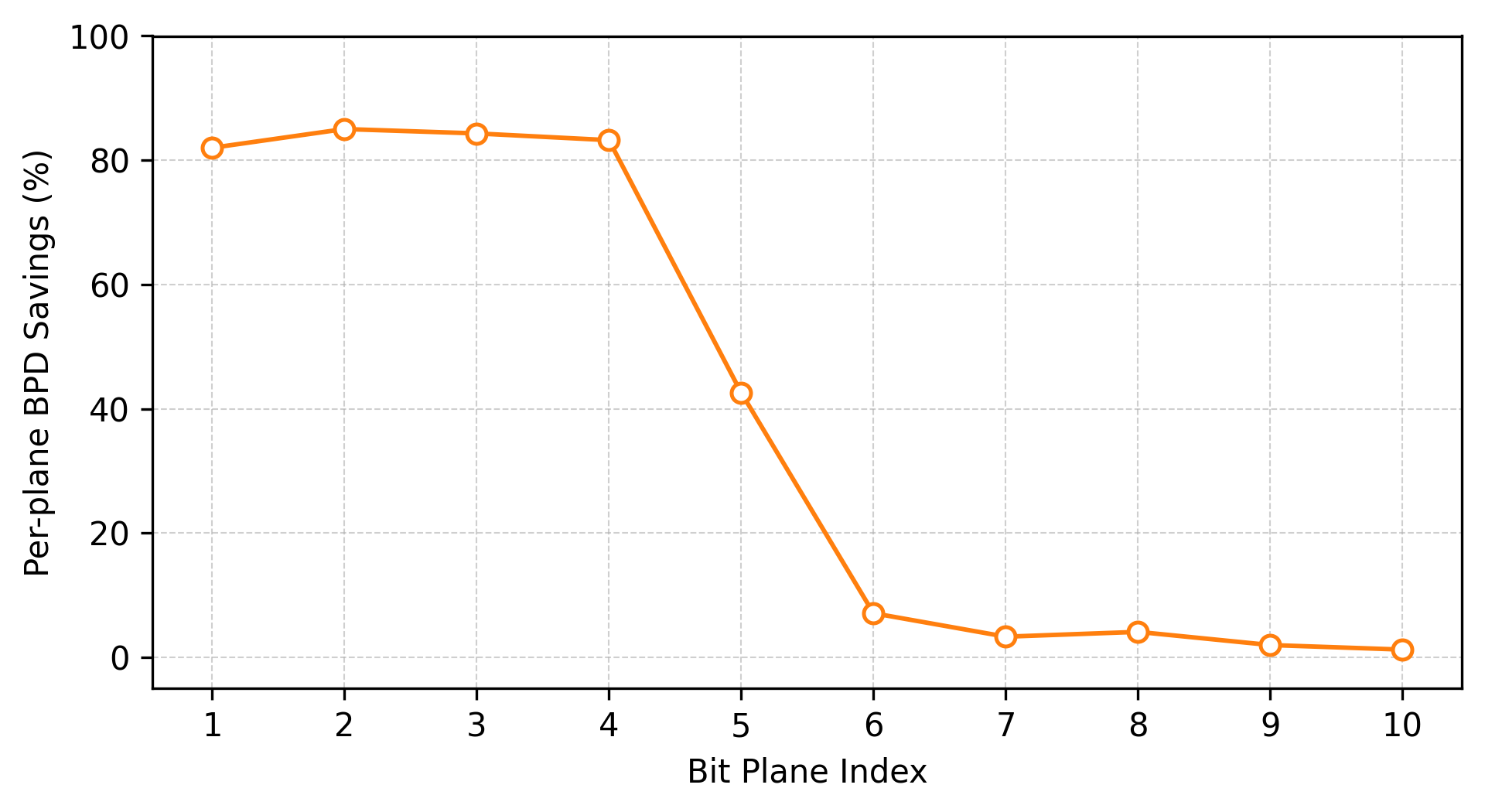}\\[-2pt]
{\scriptsize (b) WOD BPD savings}
\end{minipage}
\hfill
\begin{minipage}[t]{0.32\textwidth}
\centering
\vspace{0pt}
\includegraphics[width=\linewidth]{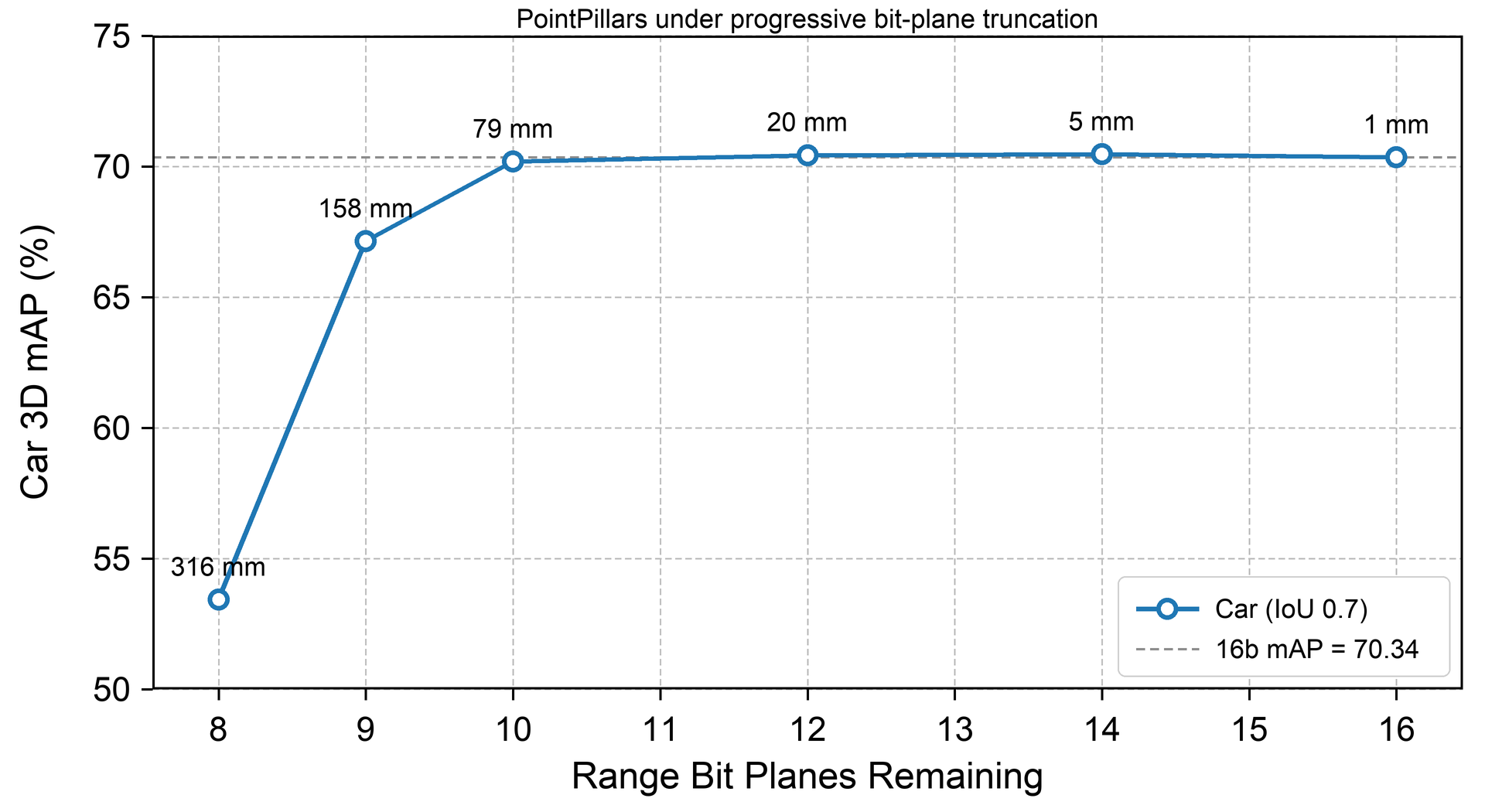}\\[-2pt]
{\scriptsize (c) Downstream $s$-sweep}
\end{minipage}
\caption{Parameter selection analysis for head depth $h$ and split index $s$. (a, b) BPD savings achieved using latent variables used to determine $h$ on SemanticKITTI and WOD. (c) Downstream 3D object detection performance under different $s$ selections, where the dashed line represents the untruncated baseline.}
\label{fig:parameter_selection}
\end{figure}

\begin{figure}[t]
\centering
\includegraphics[width=0.78\textwidth]{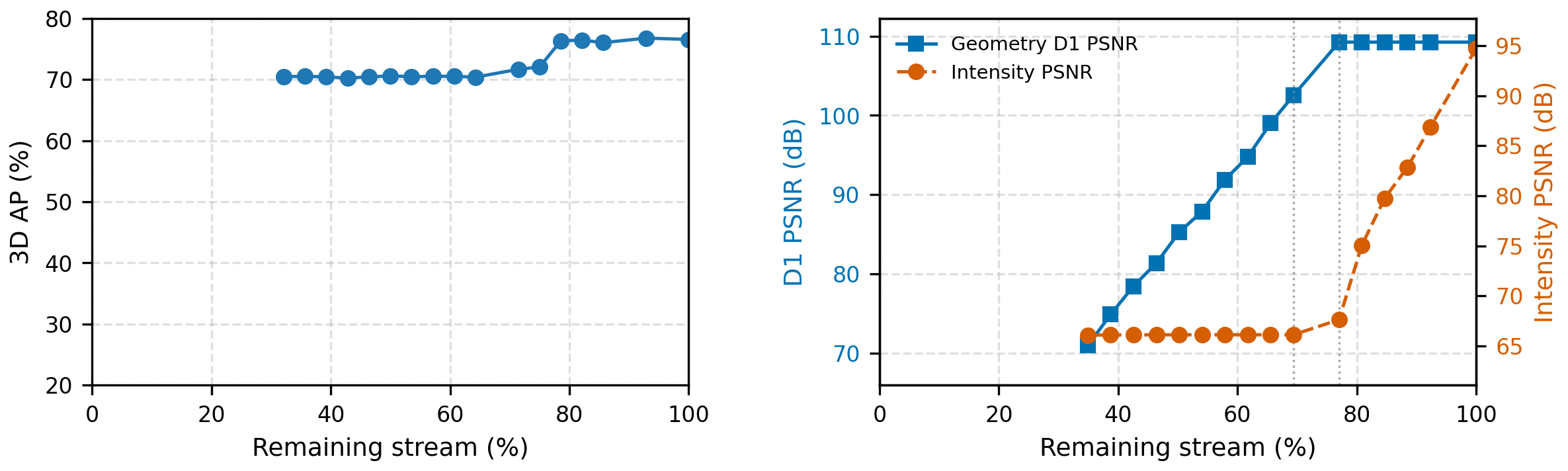}
\caption{Performance degradation under progressive bitstream truncation. Left: Downstream 3D object detection AP. Right: Geometry and attribute reconstruction quality.}
\label{fig:truncation}
\end{figure}

\begin{figure}[t]
\centering
\includegraphics[width=\textwidth]{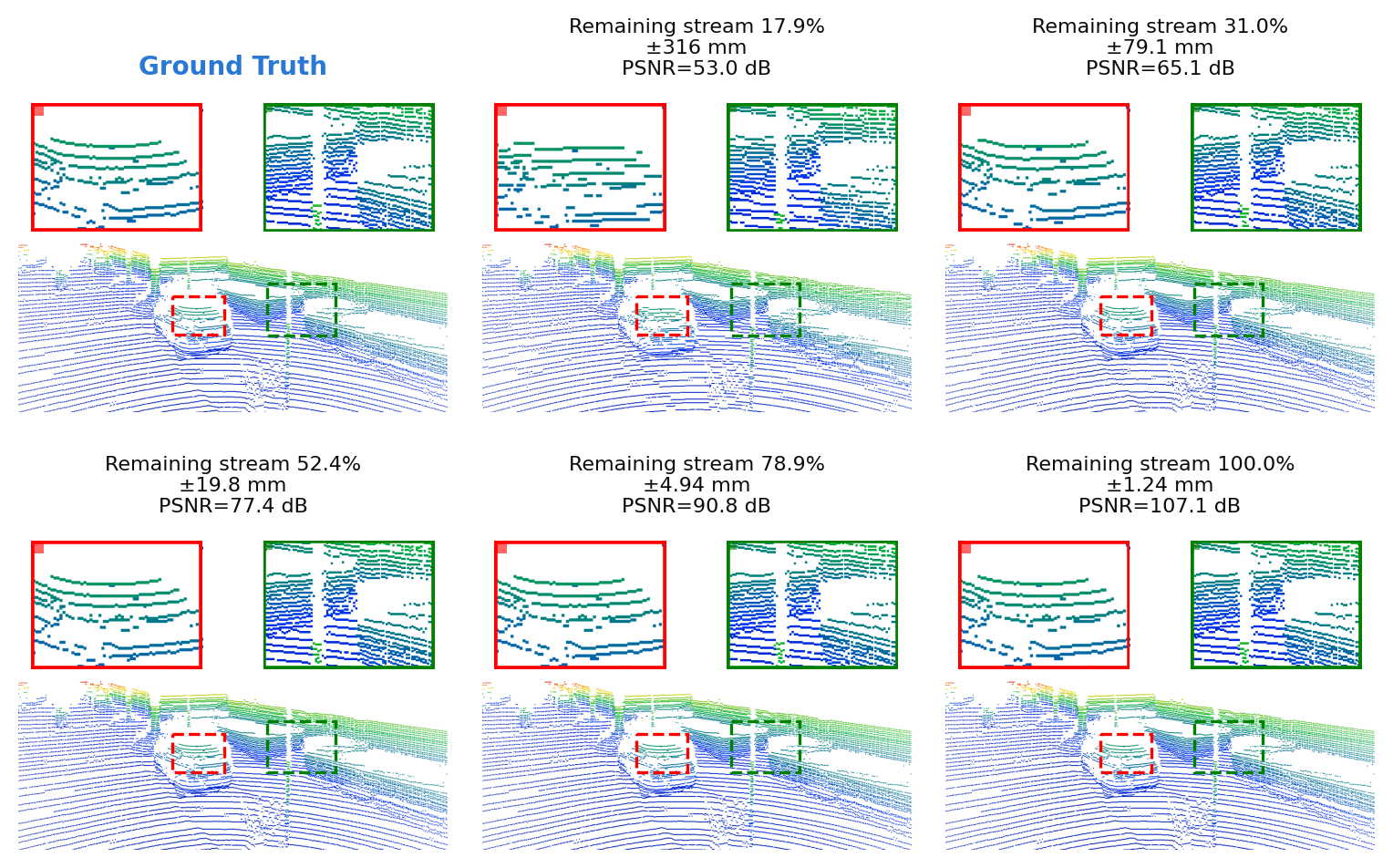}
\caption{Visualization of reconstructed LiDAR point clouds under progressive bitstream truncation.}
\label{fig:truncation_qualitative}
\end{figure}

\begin{table}[t]
\caption{BD-Rate gains to G-PCC (\%) and runtimes (in seconds). The best results are marked in bold.}
\label{tab:rd_runtime}
\centering
\scriptsize
\setlength{\tabcolsep}{1.5pt}
\begin{minipage}[t]{0.49\textwidth}
\centering
\textbf{(a) KITTI}\par\smallskip
\resizebox{\linewidth}{!}{%
\begin{tabular}{lrrrrrr}
\toprule
& \multicolumn{3}{c}{Geometry} & \multicolumn{3}{c}{Intensity} \\
\cmidrule(lr){2-4}\cmidrule(lr){5-7}
Method & BD-Rate & Enc. & Dec. & BD-Rate & Enc. & Dec. \\
\midrule
G-PCC            & 0.00 & 1.53 & 0.88
                 & 0.00 & 0.65 & 0.56 \\
Draco            & $+20.64$ & \textbf{0.02} & \textbf{0.01}
                 & $+100.05$ & \textbf{0.01} & \textbf{0.01} \\
JPEG XL          & $-11.81$ & 0.10 & 0.01
                 & $+136.28$ & 0.08 & 0.01 \\
Unicorn          & $-24.41$ & 6.65 & 6.01
                 & $-5.82$ & 3.14 & 1.99 \\
\textbf{Proteus} & $\boldsymbol{-35.41}$ & 0.16 & 0.14
                 & $\boldsymbol{-15.59}$ & 0.05 & 0.05 \\
\bottomrule
\end{tabular}%
}
\end{minipage}\hfill
\begin{minipage}[t]{0.49\textwidth}
\centering
\textbf{(b) WOD}\par\smallskip
\resizebox{\linewidth}{!}{%
\begin{tabular}{lrrrrrr}
\toprule
& \multicolumn{3}{c}{Geometry} & \multicolumn{3}{c}{Intensity} \\
\cmidrule(lr){2-4}\cmidrule(lr){5-7}
Method & BD-Rate & Enc. & Dec. & BD-Rate & Enc. & Dec. \\
\midrule
G-PCC            & 0.00 & 0.49 & 0.33
                 & 0.00 & 0.17 & 0.15 \\
Draco            & $+35.02$ & \textbf{0.03} & \textbf{0.01}
                 & $-13.41$ & \textbf{0.02} & \textbf{0.01} \\
JPEG XL          & $+2.83$ & 0.11 & 0.01
                 & $-2.72$ & 0.08 & 0.01 \\
Unicorn          & \multicolumn{6}{c}{---} \\
\textbf{Proteus} & $\boldsymbol{-42.72}$ & 0.21 & 0.20
                 & $\boldsymbol{-80.47}$ & 0.07 & 0.08 \\
\bottomrule
\end{tabular}%
}
\end{minipage}
\end{table}
\begin{figure}[t]
\centering
\includegraphics[width=\textwidth]{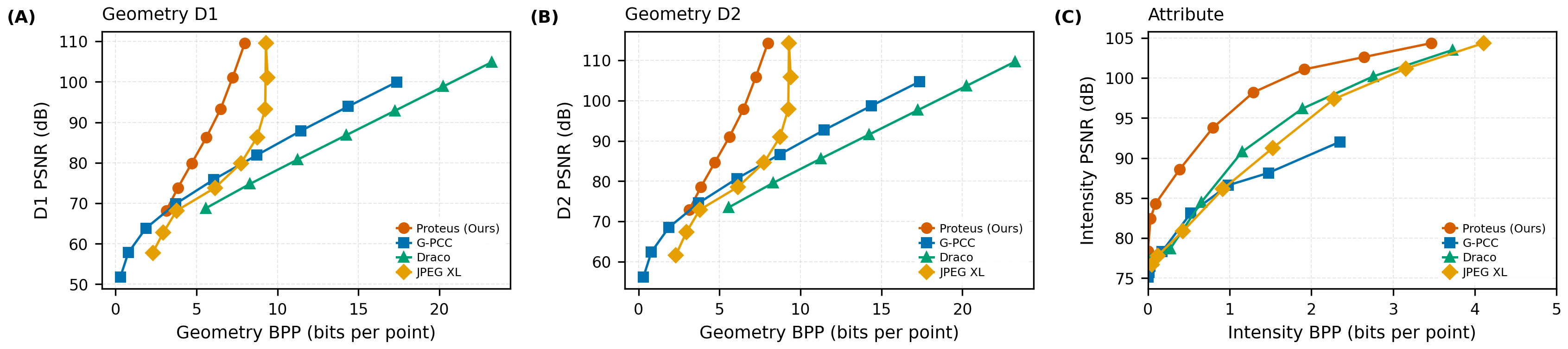}\\[-2pt]
{\small (a) WOD}\\[2pt]
\includegraphics[width=\textwidth]{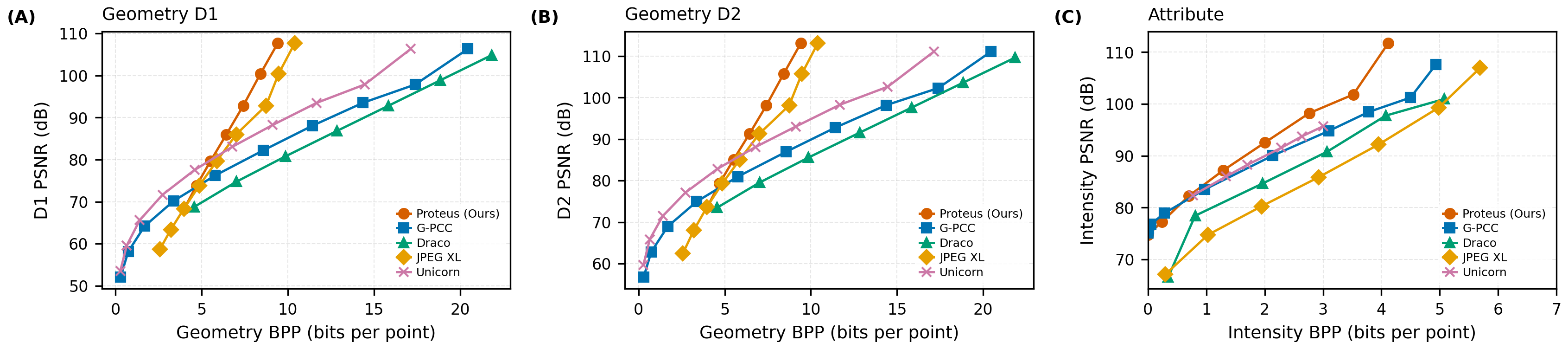}\\[-2pt]
{\small (b) SemanticKITTI}
\caption{Rate-distortion performance on WOD (top) and SemanticKITTI (bottom).}
\label{fig:rd_comparison}
\end{figure}

\begin{table}[t]
\caption{Ablation studies on (a) entropy coding variants and (b) hybrid attribute reconstruction components.}
\label{tab:ablation}
\centering
\scriptsize
\begin{minipage}[t]{0.49\textwidth}
\centering
\vspace{0pt}
\textbf{(a) Entropy coding}\\[2pt]
\begin{tabular*}{\linewidth}{@{\extracolsep{\fill}}lrc@{}}
\toprule
Variant & \shortstack{Relative\\Difference} & Robustness \\
\midrule
\text{w/o bits-back}  & +20.19\% & No \\
\text{w/o ArIB}       & +31.17\% & No \\
\text{ArIB} (coupled) & $-$6.56\% & No \\
\textbf{Proposed}       & \textbf{Ref. (0.00\%)} & \textbf{Yes} \\
\bottomrule
\end{tabular*}
\end{minipage}
\hfill
\begin{minipage}[t]{0.49\textwidth}
\centering
\vspace{0pt}
\textbf{(b) Attribute reconstruction}\\[2pt]
\setlength{\tabcolsep}{1pt}
\begin{tabular*}{\linewidth}{@{\extracolsep{\fill}}lrrr@{}}
\toprule
Variant & BD-Rate & \shortstack{Attr. Dec.\\Overhead} & \shortstack{Extra\\Storage} \\
\midrule
\text{w/o LUT and Head} & +67.8\% & +0 ms & 0 \\
\text{w/o Head}         & +19.4\% & +0.6 ms & 9 KB \\
\text{w/o LUT}          & +1.2\% & +4.6 ms & 1.3 MB \\
\textbf{Proposed}         & \textbf{Ref. (0.00\%)} & +5.2 ms & 1.3 MB \\
\bottomrule
\end{tabular*}
\end{minipage}
\end{table}

\subsection{Overview}

We build on ArIB-BPS (Section 3.3) and redesign its bitstream for truncation robustness. In ArIB-BPS the insignificant planes and the bits-back coding of the significant planes share a single coupled bitstream, and truncating its tail leaves the significant planes undecodable \citep{zhang2024learnedlossless}. Our redesign decouples them into two independent coders: SIG, a self-contained bits-back coder for the significant planes, and INS, a range coder for the insignificant planes whose first-in-first-out order tolerates truncation.

The redesign is grounded in three properties of LiDAR range images that distinguish them from generic RGB images:

\begin{itemize}
\item \textbf{A1 (bit-plane physical meaning).} The range value is a quantized physical distance. Truncating the $N$ least significant bits of every range sample halves the precision $N$ times, so bit-plane truncation is a deterministic precision degradation rather than an arbitrary distortion.
\item \textbf{A2 (attributes subordinate to range).} Attribute channels are less important than geometry in downstream perception. Industrial pipelines accordingly represent them at lower bit depth than the range channel, and we order attributes after the range in the INS bitstream.
\item \textbf{A3 (range as skeleton prior).} The high bit planes of the range predict both the low bit planes of the range and the attribute channels \citep{zhang2024learnedlossless,zhu2025serlic}. A hyperprior derived from the significant planes guides INS's encoding.
\end{itemize}

Figure~\ref{fig:overview} illustrates the resulting codec. Let $\mathbf{R} = (\mathbf{x}, \mathbf{a})$ denote the range image, where $\mathbf{x}$ is the range channel of bit depth $d$ and $\mathbf{a}$ denotes the per-pixel attribute channels (e.g., reflectance intensity). The range channel is sliced into $d$ bit planes $\mathbf{x}^{1:d}$ \citep{zhang2024learnedlossless}, partitioned into three contiguous parts: the head sub-block $\mathbf{x}^{1:h}$, the mid sub-block $\mathbf{x}^{h+1:s}$, and the tail planes $\mathbf{x}^{s+1:d}$. The significant planes ($\mathbf{x}^{1:s}$, comprising the head and mid) form SIG to perform self-contained bits-back coding, generating initial bits internally to eliminate the single-frame overhead of Section 3.2. The tail planes $\mathbf{x}^{s+1:d}$ and attributes $\mathbf{a}$ form INS, encoded via a FIFO range coder to allow progressive decoding. Parameter configurations of $h$, $s$, $d$, and attribute depths are detailed in Section~\ref{sec:experimental_setup}.

\subsection{SIG Coding}

The SIG coder encodes the significant planes $\mathbf{x}^{1:s}$ with a bits-back coding closed entirely inside SIG. SIG is split into a head sub-block $\mathbf{x}^{1:h}$ and a mid sub-block $\mathbf{x}^{h+1:s}$, where $h$ is the head depth.

Encoding proceeds in three stages \citep{townsend2019practicallossless,ryder2022splithierarchical}. (i) The mid sub-block $\mathbf{x}^{h+1:s}$ is encoded under $p(\mathbf{x}^{h+1:s})$ without latent conditioning, and the resulting bitstream provides the initial bits. (ii) The latent $\mathbf{z}$ is decoded from these bits under the posterior $q(\mathbf{z} | \mathbf{x}^{1:h})$. (iii) The head sub-block $\mathbf{x}^{1:h}$ is encoded under $p(\mathbf{x}^{1:h} | \mathbf{z})$, and $\mathbf{z}$ is encoded under $p(\mathbf{z})$. Decoding reverses this order. The net rate of SIG is

\begin{equation}
\begin{aligned}
R_{\mathrm{SIG}} ={}& -\log p(\mathbf{x}^{h+1:s})
- \log p(\mathbf{x}^{1:h} \mid \mathbf{z}) \\
& - \log p(\mathbf{z})
+ \log q(\mathbf{z} \mid \mathbf{x}^{1:h}),
\end{aligned}
\end{equation}

where the first term is the rate of the mid sub-block and the remaining three terms are the bits-back rate of the head sub-block, equal to $-\mathrm{ELBO}(\mathbf{x}^{1:h})$. The mid sub-block supplies the initial bits internally, so SIG requires no padding and does not depend on INS. The compression overhead of this decoupling design is evaluated in Section~\ref{sec:ablation}.

\subsection{INS Coding and Truncation Properties}

The INS coder encodes the tail planes of the range (the low $d - s$ bit planes) and the $d_a$-bit attribute channels $\mathbf{a}$. Following \textbf{A2}, we serialize INS as a sequence of bit-planes, ordering the range tail planes $\mathbf{x}^{s+1:d}$ before the attribute planes $\mathbf{a}^{1:d_a}$.

Let $k_r \in [s, d]$ and $k_a \in [0, d_a]$ denote the retained bit depths after truncation. A range coder provides the first-in-first-out (FIFO) encoding-decoding order that allows progressive decoding from any stream prefix, subject to the constraint that $k_a > 0$ only if $k_r = d$. To leverage the high range planes as a skeleton prior \textbf{A3}, the tail planes and attribute channels are sequentially conditioned on the reconstructed geometry. At any valid truncation point $(k_r, k_a)$, the actual rate of the INS stream is:

\begin{equation}
R_{\mathrm{INS}}(k_r, k_a)
= -\log p(\mathbf{x}^{s+1:k_r} \mid \mathbf{x}^{1:s})
- \log p(\mathbf{a}^{1:k_a} \mid \mathbf{x}).
\end{equation}

Because SIG and INS are written into independent bitstreams, truncating the INS stream leaves the base geometry intact. The decoding and reconstruction of the remaining INS components proceed as follows.

\noindent\textbf{Range Reconstruction (Precision Degradation).}
Truncating the $N = d - k_r$ least significant planes is equivalent to zero-padding, yielding the reconstructed range image $\hat{\mathbf{x}} = \mathbf{x}^{1:k_r} \cdot 2^N$. The truncation length maps directly to geometry precision through \textbf{A1}. Let $\Delta_0$ be the range precision of the full $d$-bit representation. Truncating $N$ planes halves the precision $N$ times:

\begin{equation}
\Delta_N = 2^N \cdot \Delta_0.
\label{eq:precision_degradation}
\end{equation}

For KITTI range images, $\Delta_0 \approx 1\text{mm}$ under the 16-bit quantization convention \citep{you2025renoreal}, so truncating $N=1,2,3$ planes yields approximately $2\text{mm}$, $4\text{mm}$, $8\text{mm}$ precision.

\noindent\textbf{Attribute Reconstruction (Hybrid Lossless-Predictive).}
Unlike geometry, attributes lack a strict physical-to-scale mapping and can tolerate predictive approximation. For any truncated attribute depth $k_a < d_a$, we reconstruct the remaining bits within the active quantization cell using the lossless prefix $\mathbf{a}^{1:k_a}$ and the range prior $\hat{\mathbf{x}}$:

\begin{equation}
\hat{\mathbf{a}}
= \mathbf{a}^{1:k_a}
+ \delta_{k_a} \cdot
\sigma\big(h_\theta(\hat{\mathbf{x}}, \mathbf{a}^{1:k_a}, k_a)\big),
\end{equation}

where $\delta_{k_a} = 2^{-k_a}$ is the cell width, and $\sigma(\cdot) = 0.5 + 0.5\tanh(\cdot)$ scales the prediction network $h_\theta$'s output to $[0, 1]$. The reconstruction overhead and the design choices of $h_\theta$ are evaluated in Section~\ref{sec:ablation}.

Four properties follow. (i) \textbf{No prior negotiation}: the sender and receiver need not agree on a truncation point; any INS suffix can be dropped in transit. (ii) \textbf{No geometric hallucination}: truncated range planes are not predicted, ensuring the range is reconstructed exactly at the precision the bitstream carries. (iii) \textbf{Deterministic degradation}: the geometry precision is $\Delta_N = 2^N \cdot \Delta_0$ regardless of the range content. (iv) \textbf{Progressive decoding}: INS's FIFO order allows decoding to halt at the truncation point and still recover every plane preceding it.

\subsection{Training Objective}

The training proceeds in two stages:

\noindent\textbf{Stage 1: Compression Network Training.}
We optimize the entropy model parameters using the joint rate-distortion loss under maximum capacity ($k_r=d, k_a=d_a$):

\begin{equation}
\mathcal{L}_{\mathrm{compress}}
= -\log p(\mathbf{x}^{h+1:s})
- \mathrm{ELBO}(\mathbf{x}^{1:h})
- R_{\mathrm{INS}}(d, d_a),
\end{equation}

where
\begin{equation*}
\begin{aligned}
\mathrm{ELBO}(\mathbf{x}^{1:h})
={}& \mathbb{E}_{\mathbf{z} \sim q(\mathbf{z} \mid \mathbf{x}^{1:h})}
\big[\log p(\mathbf{x}^{1:h} \mid \mathbf{z}) \\
&\qquad + \log p(\mathbf{z})
- \log q(\mathbf{z} \mid \mathbf{x}^{1:h})\big].
\end{aligned}
\end{equation*}

\noindent\textbf{Stage 2: Prediction Head Training.}
With the compression network frozen, we train the prediction head $h_\theta$. For each training instance, we sample $k_a \sim \text{Uniform}\{0, \dots, d_a\}$ to represent the progressive truncation states. The network is optimized via a cell-normalized masked Mean Squared Error (MSE) over valid range pixels ($x > 0$):

\begin{equation}
\mathcal{L}_{\mathrm{rec}}
= \frac{1}{N_{\mathrm{valid}}}
\sum_{\mathrm{valid}}
\left(
\frac{\hat{\mathbf{a}} - \mathbf{a}}{\delta_{k_a}}
\right)^2,
\end{equation}

where $\delta_{k_a} = 2^{-k_a}$. Normalizing the residual by the quantization cell width $\delta_{k_a}$ scales the errors at different truncation levels to the same relative magnitude, balancing the gradient contributions across all progressive bit-planes.

\section{Experiments}
\label{sec:experiments}

\subsection{Experimental Setup}
\label{sec:experimental_setup}

\subsubsection{Datasets.}

We conduct evaluations on the Waymo Open Dataset (WOD) \citep{sun2020waymo} and the SemanticKITTI dataset \citep{behley2019semantickitti}. WOD provides raw range images alongside the accurate emission angles of LiDAR beams \citep{zhou2022riddlelidar}. This high-fidelity angular information ensures a mathematically lossless transformation between the 2D range image grid and the 3D Cartesian point cloud coordinates. Unlike WOD, SemanticKITTI only provides raw 3D point cloud coordinates and does not release predefined beam emission angles or raw range images \citep{zhou2022riddlelidar}. Following the established evaluation protocols in prior works \citep{zhou2022riddlelidar}, our experiments on SemanticKITTI are conducted on pseudo range images projected from the 3D point clouds using estimated LiDAR beam emission angles.

\subsubsection{Baselines.}

Since no existing learned point cloud compression (PCC) frameworks support self-contained, prefix-decodable progressive coding under truncation, directly comparable baselines in this regime are unavailable. Moreover, although 2D progressive image codecs (e.g., JPEG XL \citep{sneyers2025jpegxl} or JPEG 2000 \citep{skodras2001jpeg2000}) natively support truncation, their frequency- or wavelet-domain algorithms do not translate to deterministic, bit-plane-aligned 3D coordinate degradation. We therefore select G-PCC v23 \citep{mpeg2023gpccv23}, Google Draco \citep{google2023draco}, JPEG XL \citep{sneyers2025jpegxl}, and Unicorn \citep{wang2025aversatile_attr,wang2025aversatile_geom} to evaluate and verify our compression performance under ideal channel conditions (0\% truncation).

\subsubsection{Implementation Details.}

We train separate models for WOD and SemanticKITTI. All experiments are conducted on a single NVIDIA GeForce RTX 3090 GPU using the Adam optimizer \citep{kingma2015adam} with an initial learning rate of $1\times10^{-4}$ and a step decay schedule. For our model configuration, we set the head depth $h$ to 7 (with $\text{mid}=3$) for SemanticKITTI and 6 (with $\text{mid}=4$) for WOD based on the inflection points of bits per dimension (BPD) savings achieved using latent variables in Figure~\ref{fig:parameter_selection}(a,b). To provide a necessary self-contained perceptual lower bound under progressive truncation, the split index $s$ is consistently set to 10 for both datasets based on the PointPillars evaluation \citep{lang2019pointpillars} in Figure~\ref{fig:parameter_selection}(c).

\subsection{Performance Evaluation}

\subsubsection{Truncation Robustness under Dynamic Channels.}

We evaluate the progressive degradation behavior of our codec under dynamic channel conditions by sweeping the truncation ratio of the INS stream from 0\% to 100\%. Because the SIG block utilizes the ArIB bits-back mechanism, it is non-truncatable. However, since SIG only accounts for approximately 30\% of the overall bitstream, Proteus natively supports a truncation ratio of up to approximately 70\%. Under truncation, our framework sacrifices attribute precision first to prioritize geometric preservation, as reflected by the geometry and intensity PSNR trends in Figure~\ref{fig:truncation} (right), while downstream 3D object detection AP degrades progressively down to the self-contained lower bound (Figure~\ref{fig:truncation}, left).

Specifically, the geometric error remains mathematically bounded by Equation~\ref{eq:precision_degradation}, and Figure~\ref{fig:truncation_qualitative} presents the visual results of the reconstructed point clouds under varying truncation levels.

\subsubsection{Rate-Distortion Performance under Ideal Channels.}

We evaluate the rate-distortion (R-D) performance of Proteus under ideal channel conditions (0\% truncation) using BD-Rate \citep{barman2024bdtutorial}; geometry quality follows the point-to-point metric \citep{tian2017geometrymetrics}.

\noindent\textbf{Geometry Compression.}
Proteus outperforms other baselines, achieving BD-rates of \textbf{$-35.41\%$} on SemanticKITTI and \textbf{$-42.72\%$} on WOD compared to G-PCC. This demonstrates the efficiency of our entropy coding framework. Additionally, to achieve a coordinate precision of $n$ bits, 3D voxel- or octree-based methods must subdivide the 3D space, scaling their spatial cell representation exponentially as $O(k^n)$ (where $k$ is the spatial dimension factor). This is highly efficient at low precision where sparse, empty space is skipped, but the symbol overhead scales rapidly at higher precision levels. In contrast, range-image-based methods operate on a fixed 2D grid, requiring only additional bit planes that scale the complexity linearly as $O(n)$. Consequently, while 3D methods excel at coarse reconstructions at low bitrates, range-image approaches scale more favorably and outperform them in high-precision regimes.

\noindent\textbf{Attribute Compression.}
For intensity compression, Proteus outperforms other baselines, achieving BD-rates of \textbf{$-15.59\%$} on SemanticKITTI and \textbf{$-80.47\%$} on WOD compared to G-PCC. The performance difference between the two datasets is primarily driven by variations in dataset distribution, as detailed in Appendix~\ref{app:intensity_analysis}.

\subsection{Ablation Studies}
\label{sec:ablation}

To verify the necessity of the bits-back mechanism, the ArIB scheme, and to quantify the overhead of achieving truncation robustness, we conduct an ablation study on SemanticKITTI. As shown in Table~\ref{tab:ablation}(a), removing the bits-back mechanism (\text{w/o bits-back}) increases the average rate by +20.19\%. Further removing the ArIB scheme (\text{w/o ArIB}) leads to a +31.17\% rate increase. Theoretically, \text{w/o bits-back} and \text{w/o ArIB} perform identically for single-frame encoding due to unamortized dummy bits. In practice, however, \text{w/o ArIB} requires larger protective padding to prevent ANS decoding segmentation faults, increasing the bitstream length. Finally, compared to the coupled ArIB scheme, our decoupled framework pays only a minor 6.56\% efficiency tax to enable progressive degradation, whereas ArIB entirely lacks truncation robustness.

Beyond the entropy coding framework, we also evaluate the design choices of our hybrid attribute reconstruction module. To verify the necessity of the lookup table (LUT), the CNN reconstruction head (Head), and to evaluate their complexity overhead, we conduct an ablation study on attribute prediction. As shown in Table~\ref{tab:ablation}(b), removing both components (\text{w/o LUT and Head}) increases the BD-rate by +67.8\%. Removing the CNN head (\text{w/o Head}) leads to a +19.4\% BD-rate penalty, while removing only the LUT (\text{w/o LUT}) still incurs a +1.2\% rate increase. Overall, the proposed joint LUT and Head design is computationally lightweight and economical, requiring only 1.3 MB of extra storage and introducing a minor decoding overhead of 5.2 ms.

\FloatBarrier
\section{Conclusion}
\label{sec:conclusion}

This paper presents Proteus, a learned LiDAR point cloud compression framework. By decoupling the representation into independent SIG and INS coders, Proteus achieves overall stream-level truncation robustness. Under ideal channel conditions, Proteus outperforms traditional standards and the representative learned compressor Unicorn, while natively supporting up to approximately 70\% truncation with progressive degradation.


\clearpage
\bibliographystyle{iclr2026_conference}
\bibliography{references}

\clearpage
\appendix


\begin{center}
    \bf\Large Appendix
\end{center}

\section{Evaluation Metrics}
\label{app:evaluation_metrics}

To evaluate the reconstruction quality of the LiDAR point cloud geometry and attributes, we define the mathematical formulations of the metrics below. Let $P$ denote the original point cloud and $\hat{P}$ represent the reconstructed point cloud.

\subsection{Geometric Peak Value}

To establish a normalized peak signal-to-noise ratio, the peak value $r$ is defined as the maximum nearest-neighbor distance across the entire evaluation dataset $\mathcal{D}$ \citep{tian2017geometrymetrics}:

\begin{equation}
r = \max_{P \in \mathcal{D}} \left( \max_{\mathbf{p}_i \in P} \min_{\mathbf{p}_j \in P, j \neq i} \|\mathbf{p}_i - \mathbf{p}_j\|_2 \right).
\end{equation}

Based on the dataset conventions, we set $r = 59.70$ for SemanticKITTI and $r = 57.41$ for the Waymo Open Dataset (WOD).

\subsection{Point-to-Point (D1) Metric}

The symmetric Point-to-Point Mean Squared Error ($\text{MSE}_{\text{D1}}$) measures the Euclidean distance between a point and its nearest neighbor in the corresponding point cloud \citep{tian2017geometrymetrics}:

\begin{equation}
\text{MSE}_{\text{D1}}(P, \hat{P}) = \max \left( \frac{1}{|P|} \sum_{\mathbf{p} \in P} d^2(\mathbf{p}, \hat{P}),\ \frac{1}{|\hat{P}|} \sum_{\hat{\mathbf{p}} \in \hat{P}} d^2(\hat{\mathbf{p}}, P) \right),
\end{equation}

where $d(\mathbf{p}, \hat{P}) = \min_{\hat{\mathbf{p}} \in \hat{P}} \|\mathbf{p} - \hat{\mathbf{p}}\|_2$. The geometry D1 PSNR is then calculated as:

\begin{equation}
\text{PSNR}_{\text{D1}} = 10 \log_{10} \left( \frac{3 r^2}{\text{MSE}_{\text{D1}}(P, \hat{P})} \right).
\end{equation}

\subsection{Point-to-Plane (D2) Metric}

The symmetric Point-to-Plane Mean Squared Error ($\text{MSE}_{\text{D2}}$) projects the nearest-neighbor displacement vector along the surface normal vector of the reference point \citep{tian2017geometrymetrics}:

\begin{equation}
\text{MSE}_{\text{D2}}(P, \hat{P}) = \max \left( \frac{1}{|P|} \sum_{\mathbf{p} \in P} \left( (\mathbf{p} - \hat{\mathbf{p}}_\mathbf{p}) \cdot \mathbf{n}_\mathbf{p} \right)^2,\ \frac{1}{|\hat{P}|} \sum_{\hat{\mathbf{p}} \in \hat{P}} \left( (\hat{\mathbf{p}} - \mathbf{p}_{\hat{\mathbf{p}}}) \cdot \mathbf{n}_{\hat{\mathbf{p}}} \right)^2 \right),
\end{equation}

where $\hat{\mathbf{p}}_\mathbf{p}$ represents the nearest neighbor of $\mathbf{p}$ in $\hat{P}$, and $\mathbf{n}_\mathbf{p}$ denotes the surface normal vector at point $\mathbf{p}$. The geometry D2 PSNR is formulated as:

\begin{equation}
\text{PSNR}_{\text{D2}} = 10 \log_{10} \left( \frac{3 r^2}{\text{MSE}_{\text{D2}}(P, \hat{P})} \right).
\end{equation}

\subsection{Reflectance Attribute Metric}

The attribute reconstruction accuracy is measured via the reflectance PSNR over all valid points $N_{\text{valid}}$:

\begin{equation}
\text{MSE}_{\text{ref}} = \frac{1}{N_{\text{valid}}} \sum_{i=1}^{N_{\text{valid}}} (a_i - \hat{a}_i)^2,
\end{equation}

\begin{equation}
\text{PSNR}_{\text{ref}} = 10 \log_{10} \left( \frac{V_{\max}^2}{\text{MSE}_{\text{ref}}} \right),
\end{equation}

where $a_i$ and $\hat{a}_i$ are the normalized reference and decoded reflectance values, respectively, and the peak intensity value is set to $V_{\max} = 1.0$.

\section{Proteus Architecture and LadderVAE2 State Machine}
\label{app:architecture}

Proteus implements a 2-layer hierarchical Variational Autoencoder (\text{LadderVAE2}) to execute the Decoupled Bits-back coding framework. While Proteus retains the structural layers of RangeCM's feature extraction blocks ($g_a, h_a, h_s, g_s$) to maintain architectural compatibility \citep{song2025lowlatency}, their functional roles are repurposed. Instead of projecting continuous values directly, these blocks parameterize the location parameters of discrete hierarchical distributions over binary bit-planes.

\subsection{Parameter Prediction Networks (Param Nets)}

To map the continuous feature representations to the discrete probability spaces of our bit-planes, we introduce three lightweight parameter prediction networks (heads) built with 2D convolutional layers:

\begin{enumerate}
\item \textbf{\text{q\_head\_z}:} A posterior parameter network that maps the hyper-latent feature space $z_{\text{features}}$ to the top-level location parameter $\mu_{q_z}$:
  \begin{equation}
  \mu_{q_z}
  = \text{q\_head\_z}(z_{\text{features}})
  \in \mathbb{R}^{B \times z_{\text{channels}} \times H/8 \times W/128}.
  \end{equation}

\item \textbf{\text{q\_head\_y}:} A joint posterior parameter network that concatenates the bottom-up feature extraction $y_{\text{features}}$ and the top-down reconstructed feature space $h_{s_{\text{out}}}$ to generate the bottom-level location parameter $\mu_{q_y}$:
  \begin{equation}
  \mu_{q_y}
  = \text{q\_head\_y}\big([h_{s_{\text{out}}}, y_{\text{features}}]\big)
  \in \mathbb{R}^{B \times y_{\text{dim}} \times H/4 \times W/64}.
  \end{equation}

\item \textbf{\text{p\_head\_y}:} A prior parameter network that maps the top-down features $h_{s_{\text{out}}}$ alone to the prior location parameter $\mu_{p_y}$ for bottom-level decoding:
  \begin{equation}
  \mu_{p_y}
  = \text{p\_head\_y}(h_{s_{\text{out}}})
  \in \mathbb{R}^{B \times y_{\text{dim}} \times H/4 \times W/64}.
  \end{equation}
\end{enumerate}

\subsection{Hierarchical Distribution Parameterization}

The hierarchical model comprises a top-level hyper-latent $z$ and a bottom-level main latent $y$. Both latents are modeled using Logistic distributions with a unit scale parameter.

\subsubsection{Layer $z$ (Top Level)}

\begin{itemize}
\item \textbf{Posterior $q(z|\mathbf{x})$:} where $\mu_{q_z}$ is predicted using \text{q\_head\_z}, and $z_{\text{features}} = h_a(g_a(\mathbf{x}))$.
  \begin{equation}
  q(z|\mathbf{x}) = \text{Logistic}(\mu_{q_z}, 1).
  \end{equation}

\item \textbf{Prior $p(z)$:} where the location parameter is a static zero tensor ($\mu_{p_z} = \mathbf{0}$). This design forces the prior to follow a standard standard Logistic distribution, eliminating the need for hyper-prior projection layers.
  \begin{equation}
  p(z) = \text{Logistic}(0, 1).
  \end{equation}
\end{itemize}

\subsubsection{Layer $y$ (Bottom Level)}

\begin{itemize}
\item \textbf{Posterior $q(y|\mathbf{x}, z)$:} where $\mu_{q_y}$ is generated using \text{q\_head\_y} with the concatenated features, and $h_{s_{\text{out}}} = h_s(z)$.
  \begin{equation}
  q(y|\mathbf{x}, z) = \text{Logistic}(\mu_{q_y}, 1).
  \end{equation}

\item \textbf{Prior $p(y|z)$:} where $\mu_{p_y}$ is computed from the top-down propagation using \text{p\_head\_y}.
  \begin{equation}
  p(y|z) = \text{Logistic}(\mu_{p_y}, 1).
  \end{equation}
\end{itemize}

\subsection{Bits-Back State Machine Flow}

Due to the Last-In-First-Out (LIFO) stack property of Asymmetric Numerical Systems (ANS), the compression and decompression routines must execute ``decode'' and ``encode'' operations in strictly reversed order \citep{duda2014asymmetricnumeral,townsend2019practicallossless}. Proteus utilizes the Insignificant Planes (INS) stream as the initial ANS stack state to resolve the single-frame overhead of conventional bits-back coding.

\begin{center}
\begin{tikzpicture}[
  stream/.style={
    draw,
    rounded corners=1pt,
    inner xsep=5pt,
    inner ysep=2.5pt,
    font=\footnotesize
  },
  operation/.style={
    inner xsep=1pt,
    inner ysep=1pt,
    font=\footnotesize
  },
  flow/.style={
    -{Latex[length=1.6mm,width=1.1mm]},
    line width=0.45pt
  }
]
\node[stream, anchor=west] (enc-ins) {INS Stream};
\node[font=\small\bfseries, anchor=south west, yshift=2.5mm]
  (enc-title) at (enc-ins.north west) {Encoder (Compression)};
\node[operation, right=4.5mm of enc-ins] (enc-qz)
  {(Decode $z$ via $q$)};
\node[operation, right=4.5mm of enc-qz] (enc-qy)
  {(Decode $y$ via $q$)};
\node[stream, right=4.5mm of enc-qy] (enc-sig)
  {SIG Encoded};

\node[operation, below=7mm of enc-qy] (enc-py)
  {(Recycle $y$ via $p$)};
\node[operation, left=4.5mm of enc-py] (enc-pz)
  {(Recycle $z$ via $p$)};
\node[stream, left=4.5mm of enc-pz] (enc-final)
  {Final Stream};

\draw[flow] (enc-ins) -- (enc-qz);
\draw[flow] (enc-qz) -- (enc-qy);
\draw[flow] (enc-qy) -- (enc-sig);
\draw[flow] (enc-sig.south) |- (enc-py.east);
\draw[flow] (enc-py) -- (enc-pz);
\draw[flow] (enc-pz) -- (enc-final);

\node[stream, anchor=west, yshift=-22mm] (dec-final) at (enc-ins.west)
  {Final Stream};
\node[font=\small\bfseries, anchor=south west, yshift=2.5mm]
  (dec-title) at (dec-final.north west) {Decoder (Decompression)};
\node[operation, right=4.5mm of dec-final] (dec-pz)
  {(Decode $z$ via $p$)};
\node[operation, right=4.5mm of dec-pz] (dec-py)
  {(Decode $y$ via $p$)};
\node[stream, right=4.5mm of dec-py] (dec-sig)
  {SIG Decoded};

\node[operation, below=7mm of dec-py] (dec-qy)
  {(Restore $y$ via $q$)};
\node[operation, left=4.5mm of dec-qy] (dec-qz)
  {(Restore $z$ via $q$)};
\node[stream, left=4.5mm of dec-qz] (dec-ins)
  {INS Stream};

\draw[flow] (dec-final) -- (dec-pz);
\draw[flow] (dec-pz) -- (dec-py);
\draw[flow] (dec-py) -- (dec-sig);
\draw[flow] (dec-sig.south) |- (dec-qy.east);
\draw[flow] (dec-qy) -- (dec-qz);
\draw[flow] (dec-qz) -- (dec-ins);
\end{tikzpicture}
\end{center}

\subsubsection{Compression Pipeline}

The encoding sequence is mathematically defined as follows:

\begin{enumerate}
\item \textbf{INS Initialization:} Encode the range tail planes $\mathbf{x}^{s+1:d}$ and attribute planes $\mathbf{a}^{1:d_a}$ using a range coder. The generated bitstream initializes the ANS coder (\text{MixEncoder}).

\item \textbf{Bottom-Up Feature Extraction:} Extract $y_{\text{features}} = g_a(\mathbf{x})$ and $z_{\text{features}} = h_a(y_{\text{features}})$.

\item \textbf{Decapsulate $z$ (Bits-Back Decode):} Predict posterior parameter $\mu_{q_z} = \text{q\_head\_z}(z_{\text{features}})$. \textbf{Decode} $z_{\text{ids}}$ from the active ANS stack using the CDF of $q(z|\mathbf{x})$. Reconstruct the latent representation as $z = \mu_{p_z} + \delta_z$, where $\delta_z$ is mapped from $z_{\text{ids}}$ and $\mu_{p_z} = \mathbf{0}$.
  \begin{equation}
  z_{\text{ids}} \sim q(z|\mathbf{x}).
  \end{equation}

\item \textbf{Top-Down Determinstic Step:} Generate top-down spatial features $h_{s_{\text{out}}} = h_s(z)$.

\item \textbf{Decapsulate $y$ (Bits-Back Decode):} Predict $\mu_{q_y} = \text{q\_head\_y}([h_{s_{\text{out}}}, y_{\text{features}}])$ and $\mu_{p_y} = \text{p\_head\_y}(h_{s_{\text{out}}})$. \textbf{Decode} $y_{\text{ids}}$ from the ANS stack using the CDF of $q(y|\mathbf{x}, z)$. Reconstruct the latent representation as $y = \mu_{p_y} + \delta_y$, where $\delta_y$ is mapped from $y_{\text{ids}}$.
  \begin{equation}
  y_{\text{ids}} \sim q(y|\mathbf{x}, z).
  \end{equation}

\item \textbf{SIG Synthesis:} Formulate $\hat{y} = y + h_{s_{\text{out}}}$ and generate the SIG probability distribution parameters via $g_s(\hat{y})$. Encode the significant planes $\mathbf{x}^{1:s}$ using the synthesized parameters.

\item \textbf{Recycle Latents (Prior Encode):} To satisfy LIFO, \textbf{encode} the latent variables back into the ANS stack in reverse order ($y \rightarrow z$) using their respective priors:
  \begin{itemize}
  \item First, encode the main latent $y_{\text{ids}}$ using the prior $p(y|z)$, which mathematically reduces to encoding the residual $y - \mu_{p_y}$ under $\text{Logistic}(0, 1)$.
  \item Second, encode the hyper-latent $z_{\text{ids}}$ using the prior $p(z)$, which reduces to encoding $z$ under $\text{Logistic}(0, 1)$.
  \end{itemize}
\end{enumerate}

\subsubsection{Decompression Pipeline}

At the receiver, decoding proceeds in the exact opposite direction:

\begin{enumerate}
\item \textbf{Latent Decompress (\text{\_decompress\_pz}):} Since $z$ was encoded last, it sits at the top of the ANS stack.
  \begin{itemize}
  \item \textbf{Decode} $z_{\text{ids}}$ using the prior $p(z) \sim \text{Logistic}(0, 1)$. Reconstruct $z = \mathbf{0} + \delta_z$.
  \item Propagate $z$ deterministically to obtain $h_{s_{\text{out}}} = h_s(z)$ and calculate the main prior parameter $\mu_{p_y} = \text{p\_head\_y}(h_{s_{\text{out}}})$.
  \item \textbf{Decode} $y_{\text{ids}}$ using the prior $p(y|z)$ (reconstructed from the residual under $\text{Logistic}(0, 1)$ offset by $\mu_{p_y}$). Reconstruct $y = \mu_{p_y} + \delta_y$.
  \item Formulate $\hat{y} = y + h_{s_{\text{out}}}$ and generate the conditional SIG parameters via $g_s(\hat{y})$.
  \end{itemize}

\item \textbf{Decode SIG:} Decode the significant planes $\mathbf{x}^{1:s}$ using the generated parameters.

\item \textbf{Restore Stack State (\text{\_compress\_qz}):} Before decoding the INS stream, the latent variables must be written back into the ANS stack to restore its state at the beginning of Step 1. LIFO demands the encoding order $y \rightarrow z$:
  \begin{itemize}
  \item Extract $y_{\text{features}}$ and $z_{\text{features}}$ by passing the decoded SIG planes through $g_a$ and $h_a$.
  \item Compute the posterior parameter $\mu_{q_y}$. \textbf{Encode} $y_{\text{ids}}$ using the posterior CDF of $q(y|\mathbf{x}, z)$.
  \item Compute the posterior parameter $\mu_{q_z}$. \textbf{Encode} $z_{\text{ids}}$ using the posterior CDF of $q(z|\mathbf{x})$.
  \end{itemize}

\item \textbf{Decode INS:} The ANS stack has now been returned to its initial state. Decode the insignificant range planes $\mathbf{x}^{s+1:d}$ and attribute planes $\mathbf{a}^{1:d_a}$.
\end{enumerate}

\section{Analysis of Dataset Intensity Distribution and Bit-Plane Slicing Efficiency}
\label{app:intensity_analysis}
\setcounter{figure}{0}
\renewcommand{\thefigure}{C.\arabic{figure}}
\renewcommand{\theHfigure}{C.\arabic{figure}}

The rate-distortion (R-D) efficiency of bit-plane slicing is determined by the probability density function (PDF) of the raw reflectance intensity $S \in [0, 255]$. We formalize this relationship using information-theoretic principles to demonstrate why bit-plane representation is highly suitable for WOD but less optimal for SemanticKITTI.

\begin{figure}[t]
\centering
\includegraphics[width=\textwidth]{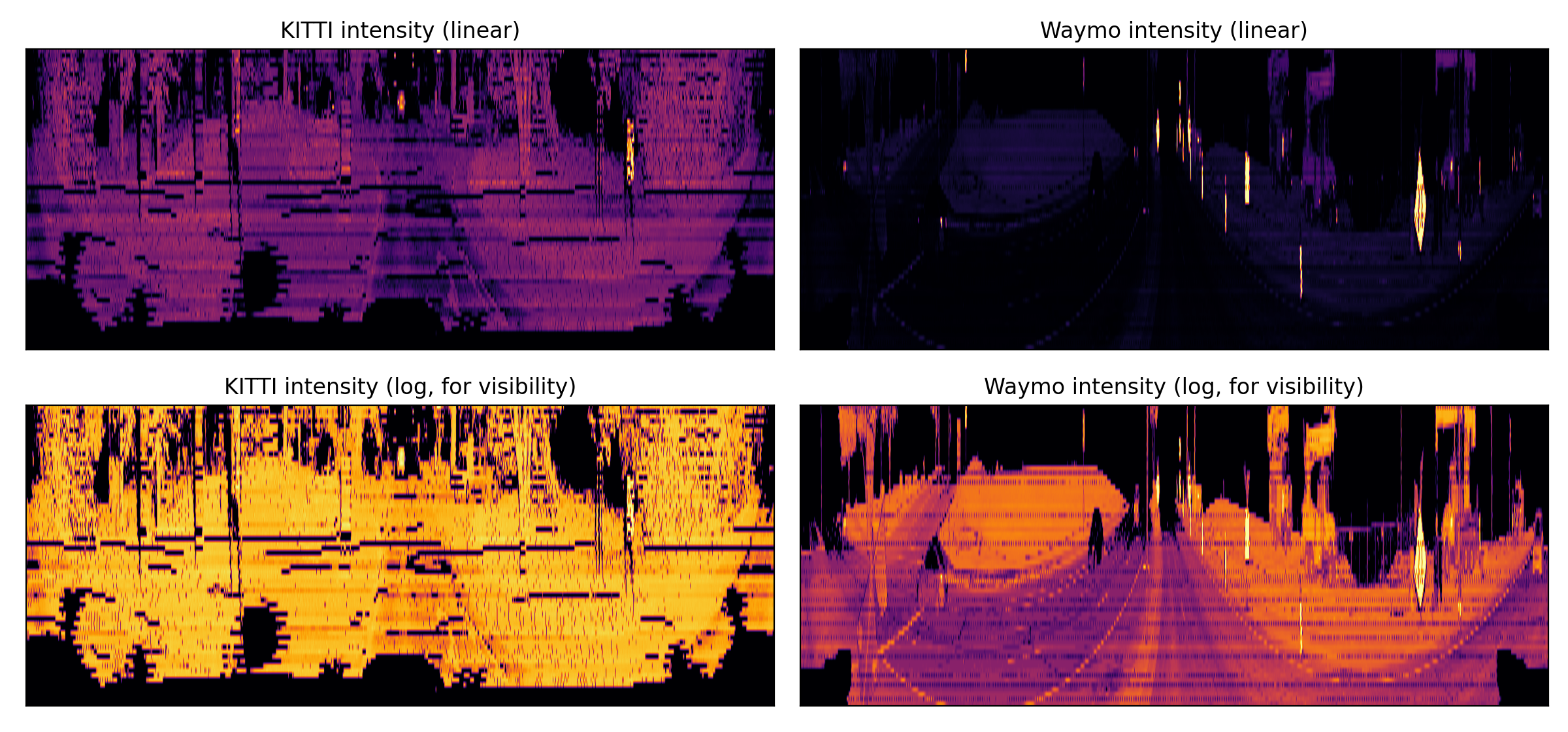}
\caption{Reflectance intensity range images under linear (top) and logarithmic (bottom) scales. Left: SemanticKITTI. Right: WOD.}
\label{fig:intensity_examples}
\end{figure}

\subsection{Mathematical Formulation of the Bit-Plane Representation}

Let $f(S)$ denote the continuous probability density function of the intensity. Bit-plane slicing decomposes the scalar value $S$ into a sequence of binary values $x^k \in \{0, 1\}$ for $k \in [1, d]$, where $d=8$. The binarization mapping for the $k$-th bit plane is defined as:

\begin{equation}
x^k
= \left\lfloor
\frac{S}{2^{d-k}}
\right\rfloor
\pmod 2.
\end{equation}

For the most significant bit-plane (MSB-0, where $k=1$), this mapping simplifies to an indicator function:

\begin{equation}
x^1
= \mathbb{I}(S \ge 128)
=
\begin{cases}
1 & \text{if } S \in [128, 255],\\
0 & \text{if } S \in [0, 127].
\end{cases}
\end{equation}

The fill rate $p_k$ of the $k$-th bit-plane is the probability $\mathbb{P}(x^k = 1)$. For $k=1$, the fill rate is determined by the cumulative density over the upper half of the intensity range:

\begin{equation}
p_1 = \int_{128}^{255} f(S) \, dS.
\end{equation}

The zero-order marginal entropy of this bit-plane is:

\begin{equation}
H(p_k)
= -p_k \log_2 p_k
- (1-p_k)\log_2(1-p_k).
\end{equation}

\begin{figure}[t]
\centering
\includegraphics[width=\textwidth]{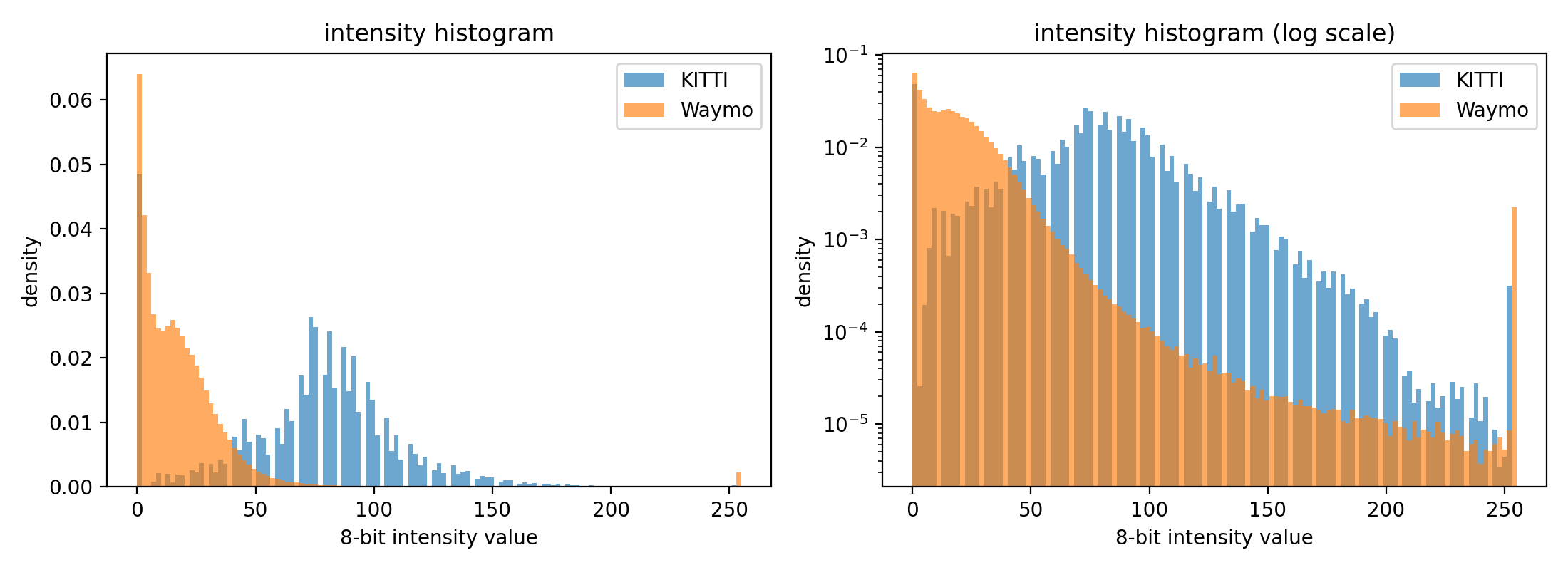}\\[-2pt]
\includegraphics[width=\textwidth]{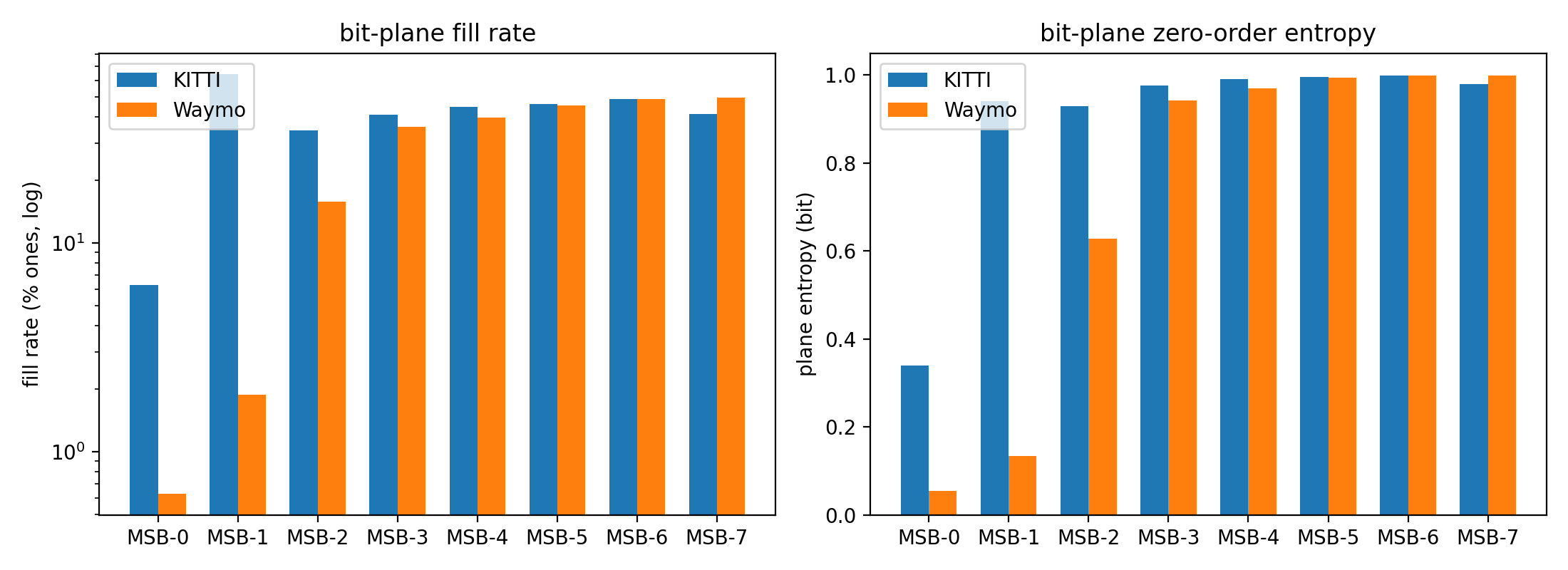}
\caption{Reflectance intensity and bit-plane statistics on SemanticKITTI and WOD. Top: Intensity histograms on linear (left) and logarithmic (right) scales. Bottom: Bit-plane fill rates (left) and zero-order entropy (right) from MSB-0 to MSB-7.}
\label{fig:intensity_statistics}
\end{figure}

\subsection{Distortion and Sparsity Properties in WOD}

As illustrated in the intensity maps and histograms (Figures~\ref{fig:intensity_examples} and~\ref{fig:intensity_statistics}), WOD's intensity distribution is highly skewed and sparse, featuring a sharp peak at zero and a heavy, power-law tail. The PDF can be approximated as a mixture model:

\begin{equation}
f_{\text{WOD}}(S)
= (1 - \epsilon) \delta(S) + \epsilon g(S),
\end{equation}

where $\delta(S)$ is the Dirac delta function representing low-intensity background surfaces, $\epsilon \approx 0.02$ represents the ratio of retroreflective objects, and $g(S)$ is a high-magnitude probability distribution spanning up to $S = 255$.

On WOD, evaluating the integral yields a very low fill rate for MSB-0 ($p_1 \approx 0.63\%$). According to the entropy formulation, the marginal coding cost is:

\begin{equation}
H_{\text{WOD}}(p_1)
\approx 0.055 \text{ bits/pixel}.
\end{equation}

However, this plane dictates the presence of the sparse, high-magnitude spikes that dominate the overall Mean Squared Error (MSE). Under an additive distortion model, the expected distortion $D$ is dominated by these uncompressed outliers if MSB-0 is truncated:

\begin{equation}
D
= \mathbb{E}[(S - \hat{S})^2]
= \int_{0}^{255}
(S - \hat{S})^2 f(S) \, dS.
\end{equation}

Because $S_i^2$ is large for retroreflective pixels, omitting $x^1$ forces $\hat{S} \le 127$, causing the squared residual $(S - \hat{S})^2$ to be exceptionally large. By coding MSB-0, these high-magnitude outliers are localized to their correct coarse cell ($S \ge 128$), reducing the expected distortion $D$ drastically. Consequently, on WOD, \textbf{transmitting MSB-0 yields an extremely high R-D efficiency (approximately $90\text{ dB/bpp}$)} because the information-theoretic cost $H(p_1)$ is near zero.

\subsection{Continuous Density Properties in SemanticKITTI}

In contrast, SemanticKITTI features a dense, continuous, mid-range distribution (Figures~\ref{fig:intensity_examples} and~\ref{fig:intensity_statistics}). Its PDF $f_{\text{KITTI}}(S)$ is characterized by a broad gaussian-like hump centered near $S \approx 85$. Evaluating the integral yields $p_1 \approx 6.0\%$, which results in a significantly higher marginal entropy:

\begin{equation}
H_{\text{KITTI}}(p_1)
\approx 0.34 \text{ bits/pixel}.
\end{equation}

For the second bit-plane (MSB-1), the fill rate on SemanticKITTI approaches $p_2 \approx 50\%$, which maximizes the entropy $H(p_2) \approx 1.0\text{ bit/pixel}$. Because the intensity field is spatially continuous and lacks extreme isolated spikes, the marginal distortion reduction of coding these high-significance planes is relatively small ($1.0\text{ dB}$ for $0.24\text{ bpp}$ on the first plane).

These mathematical properties explain the attribute compression behaviors observed in our experiments:

\begin{enumerate}
\item \textbf{WOD} natively benefits from bit-plane slicing, as the sparse, high-magnitude intensity spikes are encoded at a negligible bitrate cost.
\item \textbf{SemanticKITTI}'s continuous distribution is less suited for direct bit-plane coding, and its compression efficiency relies on the decoder-side conditional-mean reconstruction to model spatial correlations.
\end{enumerate}

\end{document}